\documentclass{article}
\usepackage[nohyperref,accepted]{icml2022_pods}
\usepackage{microtype}
\usepackage{graphicx}
\usepackage{subfigure}
\usepackage{booktabs}

\newcommand{\ft}{\texttt{FT}}
\newcommand{\lp}{\texttt{LP}}

\usepackage{amsmath}
\usepackage{amssymb}
\usepackage{mathtools}
\usepackage{amsthm}
\usepackage{wrapfig, subfig}
\usepackage{adjustbox}
\usepackage[capitalize,noabbrev]{cleveref}

\theoremstyle{plain}

\theoremstyle{definition}

\theoremstyle{remark}

\usepackage[textsize=tiny]{todonotes}
\usepackage{enumitem}

\icmltitlerunning{Exploring Adaptation Protocols for Improved Generalization and ML Safety}

\begin{document}

\twocolumn[
\icmltitle{Exploring the Design of Adaptation Protocols for Improved Generalization and Machine Learning Safety}



\icmlsetsymbol{equal}{*}

\begin{icmlauthorlist}
\icmlauthor{Puja Trivedi}{um}
\icmlauthor{Danai Koutra}{um}
\icmlauthor{Jayaraman J. Thiagarajan}{llnl}
\end{icmlauthorlist}

\icmlaffiliation{um}{Univesity of Michigan, Ann Arbor}
\icmlaffiliation{llnl}{Lawrence Livermore National Laboratory}

\icmlcorrespondingauthor{Puja Trivedi}{pujat@umich.edu}

\icmlkeywords{Machine Learning, ICML}

\vskip 0.3in
]



\printAffiliationsAndNotice{} 

\begin{abstract}
While directly fine-tuning (FT) large-scale, pretrained models on task-specific data is well-known to induce strong in-distribution task performance, recent works have demonstrated that different adaptation protocols, such as linear probing (LP) prior to FT, can improve out-of-distribution generalization. However, the design space of such adaptation protocols remains under-explored and the evaluation of such protocols has primarily focused on distribution shifts. 
Therefore, in this work, we evaluate common adaptation protocols across distributions shifts and machine learning safety metrics (e.g., anomaly detection, calibration, robustness to corruptions).
We find that protocols induce disparate trade-offs that were not apparent from prior evaluation. Further, we demonstrate that appropriate pairing of data augmentation and protocol can substantially mitigate this trade-off. Finally, we hypothesize and empirically see that using hardness-promoting augmentations during LP and then FT with augmentations may be particularly effective for trade-off mitigation.
\let\thefootnote\relax\footnote{This work was performed under the auspices of the U.S. Department of Energy by the Lawrence Livermore National Laboratory under Contract No. DE-AC52-07NA27344, Lawrence Livermore National Security, LLC.and was supported by the LLNL-LDRD Program under Project No. 22-ERD-006.}
\end{abstract}

\section{Introduction}
Through larger datasets~\cite{Yalniz19_BillionScaleSSL}, better architectures~\cite{zhai2022lit,chen2021outperform,steiner2021augreg,tolstikhin2021mixer}, and novel self-supervised learning (SSL) frameworks~\cite{He20_MoCo,Chen20_SimCLR,Grill20_BYOL,Swav_Caron20}, the quality of large-scale, pretrained models has drastically and rapidly improved; resulting in more robust \cite{Hendrycks19_SSLRobust,Liu21_SSLDatasetImbalance}, transferable \cite{Ericsson21_SSLTransfer} and semantically consistent \cite{DINO_caron21} representations. 
While directly fine-tuning (\ft) such models on task-specific data is known to improve in-distribution (ID) task performance \cite{Neyshabur,zhuang19_transferlearningsurvey,Chen20_SimCLR}, recent work finds \ft~does not effectively leverage the expressiveness of large-scale, pretrained representations and fails to match the out-of-distribution (OOD) performance of other adaptation protocols, such as the \lp~\texttt{+}~\ft~ protocol which performs linear probing (\lp) prior to \ft~\cite{Kumar22_FinetuningDistorts}.
Concurrently, Kirichenko et al.~\yrcite{Kirichenko22_LastLayerRetrain} find that simply retraining the last (classifier) layer with a small amount of ``re-weighting" or minority group data, can safeguard against spurious correlations. 
Crucially, both works suggest that well-designed adaptation protocols can improve both ID task performance and robustness. 

\begin{figure}[!t]
\centering
\includegraphics[width=0.99\columnwidth]{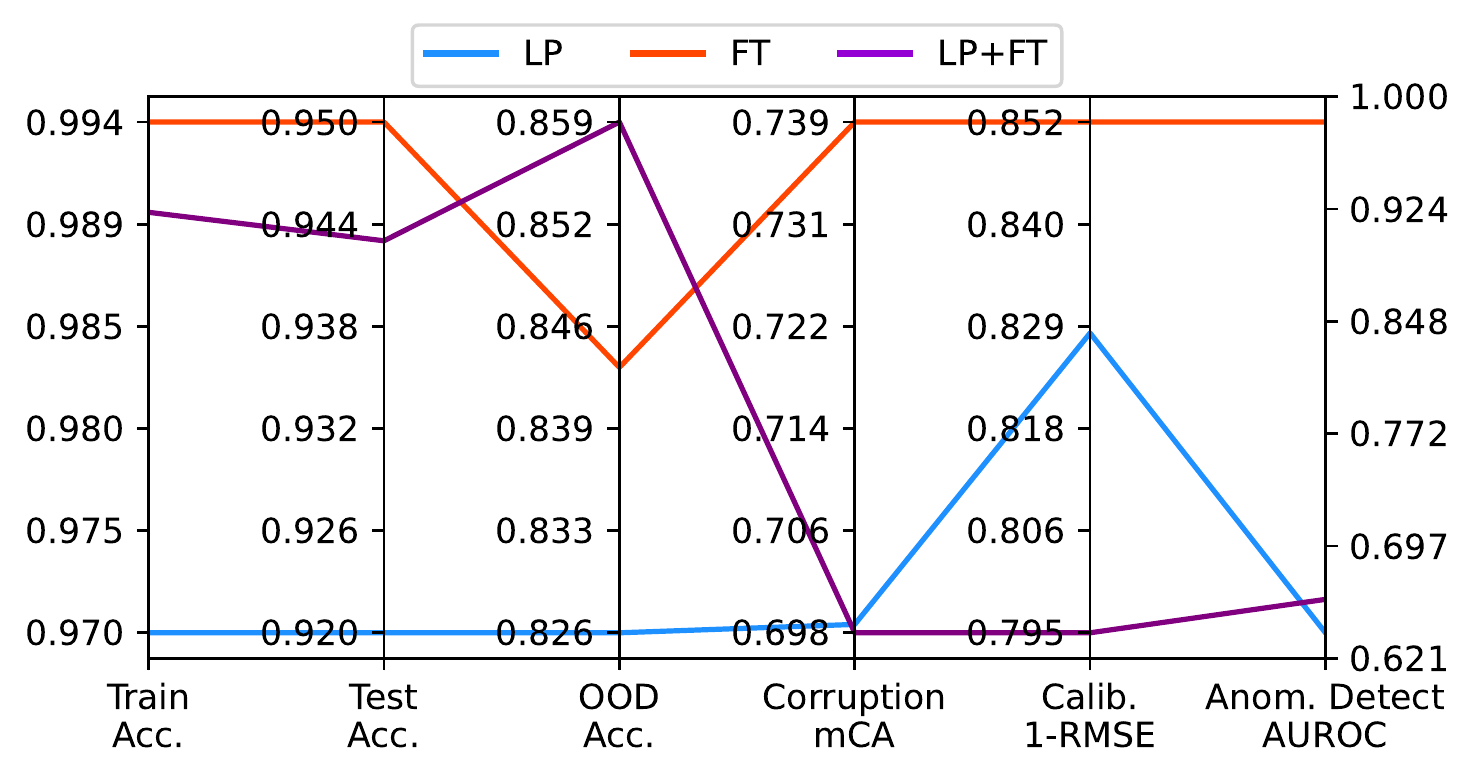}
\vspace{-0.2cm}
\caption{\textbf{Adaptation Protocols induce Trade-Offs.} While recently proposed protocol, \lp~\texttt{+}~\ft, can improve OOD accuracy, there are trade-offs with respect to other metrics. Indeed, \ft~is more effective for safety metrics.}
\label{fig:teaser}
\vspace{-0.4cm}
\end{figure}

However, practical deployment requires that models are not only robust to such shifts, but that they also perform well with respect to safety metrics, such as anomaly detection and calibration error \cite{Hendrycks21_UnsolvedProblems}. Yet, recently proposed protocols focus only on a particular aspect of generalization behavior, potentially to the detriment of others. For example, while the \lp~\texttt{+}~\ft~protocol improves OOD accuracy, its performance lags behind \ft~on other metrics (see Fig. \ref{fig:teaser}). Understanding and mitigating this trade-off is critical as all aspects are important to high-impact, low data tasks, such as healthcare applications. 

Diversity-promoting data augmentation, such as RandAug \cite{Cubuk20_RandAug}, and CutMix \cite{Yun19_CutMix}, are becoming the \textit{de facto} approach to improve model generalization. However, when not designed carefully, such sophisticated augmentations can adversely impact safety metric performance \cite{Chun20_EmpircalEval}. In practice, it is unknown what characteristics of augmentations are beneficial to adaptation and where augmentations should be incorporated into adaptation protocols to maximize their benefits. Indeed, as shown in Fig. \ref{fig:id_vs_ood_aug}, na\"ively incorporating such augmentations into adaptation protocols can lead to poorer performance than both \lp~\texttt{+}~\ft~and simple \ft. Therefore, in this paper, we holistically evaluate the behavior of adaptation protocols under distribution shifts as well as with respect to various safety metrics, and investigate how augmentations can be effectively leveraged to improve adaptation behavior.

\textbf{Proposed Work.} We first make an important finding: while state-of-the-art adaptation protocol, \lp~\texttt{+}~\ft, improves OOD accuracy, it trails behind simple \ft~on safety metrics. This emphasizes the need for a holistic evaluation when understanding the behavior of task-specific adaptation strategies. We then use this evaluation to explore how augmentations influence the  generalization behavior of different adaptation protocols. Given insights from this study, we hypothesize that hardness-promoting augmentations are needed during \lp, while  diversity-promoting augmentations can be used during \ft, for effectively implementing \lp~\texttt{+}~\ft~in practice. We verify this by employing virtual-adversarial training \cite{Miyato17_Vat} during \lp~and demonstrate significant improvements.

\begin{itemize}[noitemsep,nolistsep]
    \item\noindent \textbf{Holistic Evaluation.} We evaluate all protocols with respect to in-distribution accuracy, out-of-distribution accuracy, calibration error, anomaly detection performance and robustness to corruptions.
    
    \item\noindent \textbf{The Effect of Augmentations in Adaptation.} We show that incorporating augmentations at different stages of adaptation protocols can lead to disparate generalization performance.
    
    \item\noindent \textbf{Hardness-Based Augmentations Improve Performance.} By using virtual-adversarial training during \lp~to promote a more amenable initialization for subsequent \ft, we are able to improve generalization behavior and performance on safety metrics.
\end{itemize}
\section{Background}\label{sec:related_work}
In this section, we briefly discuss recently proposed adaptation protocols, augmentation strategies and safety metrics.

\textbf{Adaptation Protocols.} In addition to direct \ft~on downstream task data or simply training a new classifier through \lp, additional adaptation protocols have recently proposed to improve the robustness of adapted models. Kumar et al.~\yrcite{Kumar22_FinetuningDistorts} argue that large-scale, pretrained models have sufficiently expressive features to perform well on both ID and OOD data. However, direct \ft~can distort pretrained features toward ID data, harming OOD performance. To mitigate this distortion, they propose  \lp~prior to \ft~(abbrev. \lp~\texttt{+}~\ft) and find this protocol improves OOD performance. Concurrently, Kirichenko et al.~\yrcite{Kirichenko22_LastLayerRetrain} find that retraining the last-layer of a model on minority group or re-weighting data can significantly improve robustness to spurious correlations. Like Kumar et al.~\yrcite{Kumar22_FinetuningDistorts}, they argue that the model has learned expressive features but these features are being poorly utilized, e.g., the classifier relies upon spurious instead of core features. Here, we focus on the \lp~\texttt{+}~\ft~protocol as it is effective, inexpensive and re-weighting data is not available. 
We use adaptation, instead of transfer, to emphasize that we desire strong performance across safety measures in addition to strong downstream task performance.

\textbf{Data Augmentation.} Instead of building larger models or obtaining more data, data augmentation has been shown to be highly effective at improving the robustness, and generalization of models. We focus on popular, effective strategies including: AugMix \cite{Hendrycks20_AugMix}, AutoAug \cite{Cubuk18_AutoAugment}, CutMix \cite{Yun19_CutMix}, CutOut \cite{Devries17_cutout}, MixUp \cite{zhang18_mixup} and RandAug \cite{Cubuk20_RandAug}. 

\textbf{Machine Learning Safety.} Safe deployment of ML models requires that models are robust and reliable. While there are several aspects of model safety including robustness to distribution shift and adversarial samples \cite{Hendrycks21_UnsolvedProblems}, we focus on how well models are able to classify corrupted images \cite{Hendrycks19_CIFAR10C}, how well calibrated uncertainty estimates are \cite{Guo17_Calibration} and how well anomalous samples can be detected \cite{Hendrycks17_BaselineOOD, Hendrycks19_OE}. Evaluating additional aspects of ML safety is left to future work.
\section{Designing Adaptation Protocols}
In this section, we investigate the behavior of three adaptation protocols, with respect to both OOD generalization and ML safety metrics. We then investigate how incorporating popular diversity-promoting augmentations into these protocols impacts performance. Finally, we find that hardness-promoting augmentation can improve performance across all metrics. We first introduce the experimental setup.

\textbf{Experimental Set-up.} A ResNet-50 MoCoV2 \cite{He20_MoCo} model pretrained on ImageNet-1K is used as the base-feature extractor to ensure high quality, expressive representations. CIFAR-10 is the ID adaption dataset, while STL10 is the OOD dataset for which strong performance is also desired. Mean corruption accuracy (mCA) on CIFAR-10-C \cite{Hendrycks19_CIFAR10C}, RMS calibration error, and AUROC when detecting anomalous inputs are the considered safety metrics \cite{Hendrycks17_BaselineOOD}. mCA is the model's accuracy over 15 different corruptions and 5 different severities. Calibration error is measured as: $\sqrt{\mathbb{E}_{C}\left[(\mathbb{P}(Y=\hat{Y} \mid C=c)-c)^{2}\right]}$, where $C$ is confidence, $\hat{Y}$ is the model's prediction, and  $Y$ is the ground-truth label. Samples from the Blobs, Gaussian, LSUN, Places69, 
Rademacher, Textures, and SVHN datasets are considered anomalous. Our evaluation protocol closely follows 
\citeauthor{Hendrycks21_PixMix} For the \lp~protocol, we train only the classifier for 200 epochs with LR=30. For \ft, the entire model is trained for 20 epochs with LR=1e-5. For \lp~\texttt{+}~\ft, the model's classifier is initialized with the solution found by \lp, and then it is fine-tuned for 20 epochs. A grid-search was conducted to determine the LR for \lp~and \ft. When using augmented protocols, the same LRs are used. Note, all results were obtained by averaging over 3 seeds \footnote{Code to be released at: \url{https://github.com/pujacomputes/classifier_playground}}. 

\textbf{Need for Holistic Evaluation.}
As discussed in Sec. \ref{sec:related_work}, Kumar et al.~\yrcite{Kumar22_FinetuningDistorts} propose \lp~prior to \ft~as a means of mitigating feature distortion and improving OOD accuracy. In Table. \ref{table:base_protocols}, we indeed see that \lp~\texttt{+}~\ft~has better OOD accuracy than both \lp~and \ft. However, \lp~\texttt{+}~\ft's performance lags behind \ft's on robustness to corruptions, calibration, and anomaly detection. This indicates that while mitigating feature distortion is important to ensure that \ft~does not over-fit to the ID task, additional distortion may in fact be necessary for improving safety performance. Therefore, we ask how to modify the existing \lp~\texttt{+}~\ft~protocol such that pretrained features are distorted in a way that is amenable to both improved OOD accuracy and safety.  

\begin{table}[h]
 \centering
 \caption{\small{\textbf{Protocol Performance vs. Safety.} }Best performance is shown in bold. Second best is underlined. \ft~outperforms \lp~\texttt{+}~\ft~on all metrics but OOD Accuracy.}\label{table:base_protocols}
 \vspace{-0.3cm}
 \begin{adjustbox}{max width=\columnwidth}
 \begin{tabular}{l r r r r r} 
\toprule
    \cmidrule(r){1-6}
  \textbf{Protocol}  & \textbf{mCA} & \textbf{RMSE} $\downarrow$ & \textbf{AUROC}  & \textbf{ID Acc.}  & \textbf{OOD Acc.}   \\
    \midrule
  \lp~   & \underline{0.6909} & \underline{0.1697} & 0.6206  & 0.9139 & 0.8194 \\
  \ft~   & \textbf{0.7468} & \textbf{0.1366} & \textbf{1.0} & \textbf{0.9558}  & \underline{0.8434} \\
  \lp~\texttt{+}~\ft~& 0.69 & 0.2166 & \underline{0.6454} & \underline{0.9460} & \textbf{0.8656}  \\ 
\bottomrule
\end{tabular}
\end{adjustbox}
\end{table}

\begin{figure}[!t]
\centering
\includegraphics[width=1.0\columnwidth]{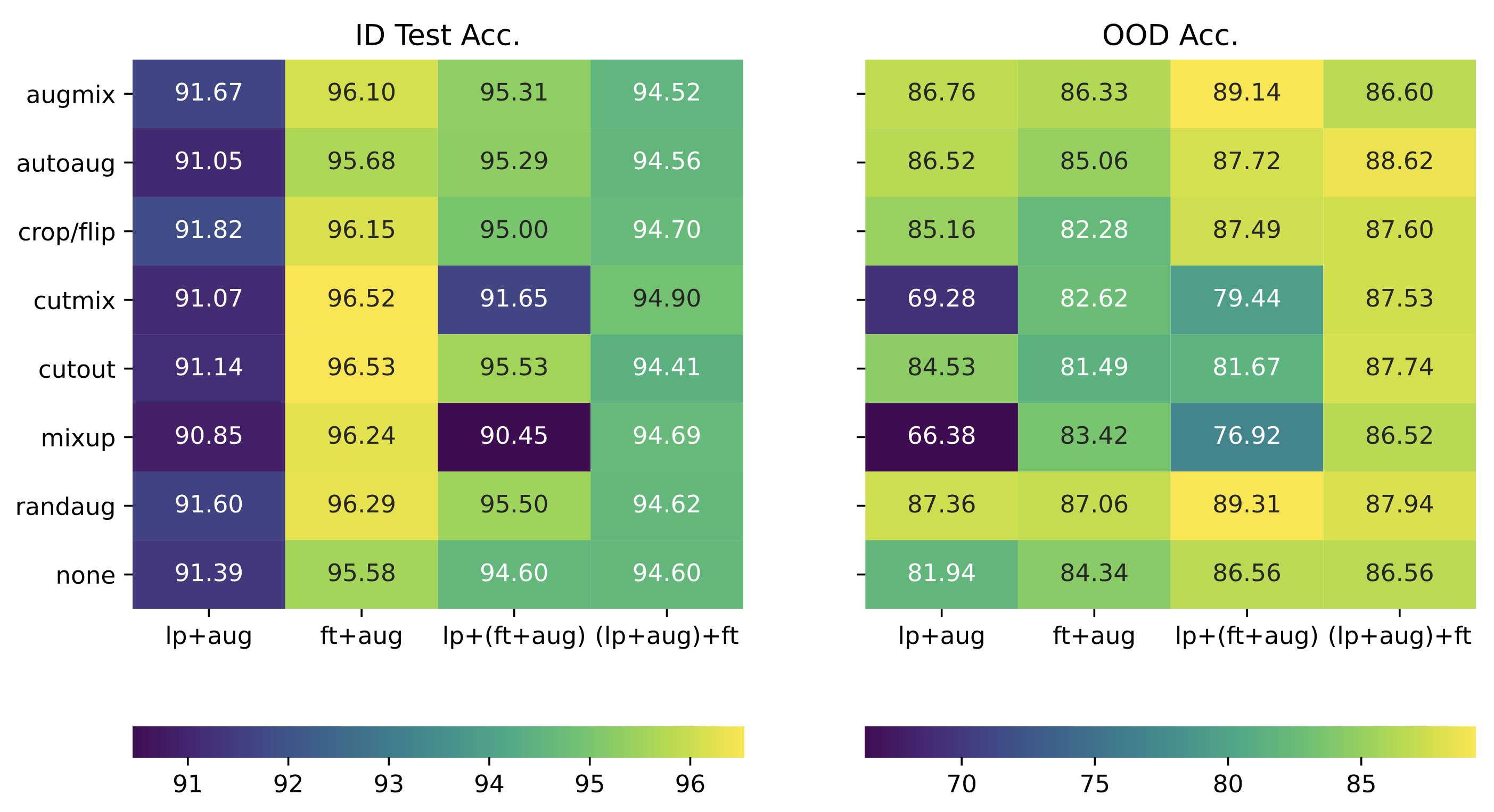}
\vspace{-0.2cm}
\caption{\textbf{Adding Augmentations to Protocols.} Most augmentations improve ID and OOD performance. However, naively adding MixUp and CutMix can be harmful.}
\label{fig:id_vs_ood_aug}
\vspace{-0.4cm}
\end{figure}

\textbf{Role of Augmentations.}
Data augmentation is well-known to be effective in improving the robustness and generalization of end-to-end training \cite{Hendrycks20_AugMix,Chun20_EmpircalEval}. However, relatively less work has focused on the role data augmentation plays when adapting high-quality pretrained representations to downstream tasks. Here, we evaluate 7 different diversity promoting augmentation strategies, including AugMix \cite{Hendrycks20_AugMix}, AutoAug \cite{Cubuk18_AutoAugment}, RandCrop+RandFlip, CutMix \cite{Yun19_CutMix}, CutOut \cite{Devries17_cutout}, MixUp \cite{zhang18_mixup}, and RandAug \cite{Cubuk20_RandAug}), as they are applied at different points of adaptation protocols. Namely, \lp~\texttt{+}aug, \ft\texttt{+}aug, during \lp~and not during \ft, \textit{i.e.}, (\lp \texttt{+}aug)~\texttt{+}~\ft, and vice-versa \lp~\texttt{+}~(\ft\texttt{+}aug). We begin by determining how these augmented protocols effect ID vs OOD performance and make the following observations from Fig. \ref{fig:id_vs_ood_aug}

\textit{Soft-Cross Entropy Loss Distorts Features.} CutMix and MixUp interpolate between samples and labels during training. Therefore, these strategies require models to be trained with the soft cross entropy loss. Noticeably, under the \lp~\texttt{+}~(\ft\texttt{+}aug) protocol, with CutMix or MixUp, models have worse ID and OOD performance than \lp~\texttt{+}~\ft, without augmentations. Similarly, \lp~\texttt{+}aug and \ft~\texttt{+}aug protocols have poorer OOD performance than their augmentation-free counterparts. 

\textit{Using Augmentations improves OOD and ID Accuracy.}
Across all protocols and all augmentations (except for MixUp and CutMix), we see that OOD accuracy and often ID accuracy are improved relative to analogous augmentation-free protocols. Indeed, we particularly observe that, incorporating augmentations with \lp~\texttt{+}~(\ft~\texttt{+} aug), can substantially improve OOD performance with comparable ID performance. 
For example, RandAug and AugMix both achieve $\geq89\%$ OOD accuracy, in comparison to plain \lp~\texttt{+}~\ft's $86.56\%$.

\textit{Fine-tuning without augmentations can recover from poor \lp~solutions.} While the OOD accuracy for CutMix and MixUp deteriorates across almost all protocols, (\lp~\texttt{+} aug)~\texttt{+}~\ft~is a notable exception. Here, we see that even if the classifier is poorly initialized after \lp, \ft~without any augmentations is still able to recover strong ID and OOD performance. Indeed, even when using MixUp/CutMix during \lp~, (\lp~\texttt{+}aug)~\texttt{+}~\ft~still outperforms plain \lp~\texttt{+}~\ft's OOD accuracy.

In summary, these observations suggest that diversity promoting, pixel-level augmentations are more effective at mitigating feature distortion to improve OOD accuracy and \ft~after \lp~is robust to poor \lp~initializations. Furthermore, we find that when taking into account the aforementioned holistic evaluation, incorporating augmentations can also improve safety performance. For brevity, we report safety performance for RandAug and Augmix, the two best performing augmentation strategies, as well as CutMix, a poorly performing strategy.

\begin{table}[h]
 \centering
 \caption{\small{\textbf{Protocol with Augmentation Performance vs. Safety.} }Results shown for AugMix, RandAug and CutMix due to space constraints. Best performance in bold. Second best are underlined.}\label{table:proto_safety}
 \vspace{-0.3cm}
 \begin{adjustbox}{max width=\columnwidth}
 \begin{tabular}{l r r r r r} 
\toprule
    \cmidrule(r){1-6}
  \textbf{Protocol}  & \textbf{mCA} & \textbf{RMSE} $\downarrow$ & \textbf{AUROC}  & \textbf{ID Acc.}  & \textbf{OOD Acc.}   \\
    \midrule
\lp      & 0.6909 & 0.1697 & 0.6206 & 0.9139 & 0.8194 \\
\lp\texttt{+}augmix      & 0.7264 & 0.1312 & 0.6477 & 0.9167 & 0.8676 \\
\lp\texttt{+}cutmix      & 0.6891 & 0.1333 & 0.5397 & 0.9107 & 0.6928 \\
\lp\texttt{+}randaug     & 0.7126 & 0.1259 & 0.6357 & 0.9160 & \underline{0.8736} \\
\midrule
\ft      & 0.7468 & 0.1367 & \textbf{1.0000} & \underline{0.9558} & 0.8438 \\
\ft\texttt{+}augmix     & \textbf{0.8139} & \textbf{0.0890} & \textbf{1.0000} & \textbf{0.9610} & 0.8632 \\
\ft\texttt{+}cutmix      & 0.7669 & 0.1345 & \textbf{1.0000} & \textbf{0.9652} & 0.8261 \\
\ft\texttt{+}randaug     & \underline{0.7871} & \textbf{0.0824} & \textbf{1.0000} & \textbf{0.9629} & \underline{0.8706} \\
\midrule
\lp~\texttt{+}~\ft    & 0.6900 & 0.2166 & 0.6455 & 0.9460 & 0.8655 \\
\lp~\texttt{+}~(\ft\texttt{+}augmix)  & \underline{0.7829} & \underline{0.1089} & 0.8074 & \underline{0.9531} & \textbf{0.8914} \\
\lp~\texttt{+}~(\ft\texttt{+}cutmix)  & 0.6663 & 0.1477 & 0.2677 & 0.9165 & 0.7944 \\
\lp~\texttt{+}~(\ft\texttt{+}randaug) & 0.7714 & 0.1190 & \underline{0.8305} & \underline{0.9550} & \textbf{0.8931} \\ 
\midrule
(\lp\texttt{+}augmix)~\texttt{+}~\ft~ & 0.7136 & 0.1856 & 0.5533 & 0.9451 & 0.8660 \\
(\lp\texttt{+}cutmix)~\texttt{+}~\ft~ & 0.6926 & 0.1724 & 0.8062 & 0.9490 & \underline{0.8753} \\
(\lp\texttt{+}randaug)~\texttt{+}~\ft~& 0.7126 & 0.1259 & 0.6357 & 0.9462 & \underline{0.8794} \\
\bottomrule
\end{tabular}
\end{adjustbox}
\end{table}

\textbf{Augmented Protocols Improve Safety.}
Across all protocols, we see that RandAug and AugMix improve the safety performance in comparison to not using any augmentation. CutMix, which significantly harmed OOD accuracy under most protocols, occasionally provides some improved calibration or corruption performance. Notably, \ft, which already had strong performance without augmentations, provides further improvement, while \lp~\texttt{+}~(\ft\texttt{+}aug) has the second best performance across safety metrics and the best OOD accuracy.

\textbf{Hardness Promoting Augmentations.}
Our results in Table. \ref{table:proto_safety} suggest that modifying the \lp~step prior to \ft~may be a viable strategy for simultaneously improving both OOD and safety performance. In particular, we hypothesize that hardness-promoting augmentations should be used during \lp~and diversity-promoting augmentations should be used during \ft. Hardness-promoting augmentations will ensure that a smooth and robust classifier is learnt during \lp, which should improve OOD performance during the subsequent fine-tuning step. While we leave theoretical analysis of our hypothesis to future work, we empirically verify it by using virtual adversarial training (VAT) \cite{Miyato17_Vat} during \lp~ to initialize the classifier prior to \ft. 

In a nutshell, VAT enforces local distribution smoothness by minimizing the KL-divergence between the predictions of perturbed pairs of examples, where the samples are adversarially perturbed such that outputs differ after perturbation. By training on such hard samples, classifiers become more robust and locally smooth. Moreover, because we are only applying VAT to the penultimate layer's representation (during \lp), this step remains relatively inexpensive.

\begin{table}[h]
 \centering
 \caption{\small{\textbf{Benefits of Hardness Promoting Augmentations.}} Incorporating VAT into the \lp~step leads to further improvements over previously identified high performing protocols. }\label{table:proto_hard}
 \vspace{-0.3cm}
 \begin{adjustbox}{max width=\columnwidth}
 \begin{tabular}{l r r r r r} 
\toprule
    \cmidrule(r){1-6}
  \textbf{Protocol}  & \textbf{mCA} & \textbf{RMSE} $\downarrow$ & \textbf{AUROC}  & \textbf{ID Acc.}  & \textbf{OOD Acc.}   \\
    \midrule
\ft\texttt{+}augmix     & \textbf{0.8139} & \textbf{0.0890} & \textbf{1.0000} & \textbf{0.9610} & 0.8632 \\
\ft\texttt{+}randaug     & 0.7871 & \textbf{0.0824} & \textbf{1.0000} & \textbf{0.9629} & 0.8706 \\
\midrule
\lp~\texttt{+}~(\ft\texttt{+}augmix)  & 0.7829 & 0.1089 & 0.8074 & \underline{0.9531} & 0.8914 \\
\lp~\texttt{+}~(\ft\texttt{+}randaug) & 0.7714 & 0.1190 & 0.8305 & \underline{0.9550} & 0.8931 \\ 
\midrule
(\lp\texttt{+}vat)~\texttt{+}~\ft          & 0.7442       & 0.1645        & 0.871        & \textbf{0.9611}       & 0.8909 \\
(\lp\texttt{+}vat)~\texttt{+}~(\ft\texttt{+}augmix)  & \textbf{0.8135}       & \textbf{0.0817}& 0.9253 & \textbf{0.9638} & \underline{0.9132} \\
(\lp\texttt{+}vat)~\texttt{+}~(\ft\texttt{+}randaug) & \underline{0.8006} & \underline{0.0900} & \underline{0.9467} & \textbf{0.9655}       & \textbf{0.9219} \\
\bottomrule
\end{tabular}
\end{adjustbox}
\end{table}

As shown in Table. \ref{table:proto_hard}, we find that training on such examples leads to significant improvements across all-metrics relative to the best performing augmented protocols. Indeed, by incorporating VAT during \lp, we are able to surpass the best OOD accuracy while performing comparably to the \ft\texttt{+}aug protocol in terms of the safety metrics. Overall, the proposed (\lp\texttt{+}vat)~\texttt{+}~(\ft\texttt{+}aug) is a viable strategy for improving performance with respect to both distributional shifts and safety measures.
\section{Conclusion and Future Directions}
In this work, we explored how modifications to common adaptation protocols influence generalization under distribution shifts and performance with respect to various ML safety metrics. We make the somewhat surprising finding that while \lp~\texttt{+}~\ft~does achieve impressive OOD accuracy, simple \ft~outperforms on the safety scores. We then find that diversity-inducing, pixel-based augmentations can circumvent this challenge to an extent. However, to jointly achieve the benefits of \ft~and \lp~\texttt{+}~\ft, hardness-inducing augmentation strategies, such as VAT, which generate challenging input perturbations, are critical. Indeed, doing so begins to close the gap with \ft~on safety metrics, while surpassing the best OOD accuracy achieved by other protocols. There are several interesting directions for future work:  

\textit{Expanded Evaluation.} We would like to expand our analysis to include adversarial robustness and larger datasets. Moreover, we are also interested in using representational analysis tools such as prediction depth \cite{Baldock21_ExampleDifficulty} or similarity metrics \cite{Kornblith19_CKA, Raghu17_SVCCA} to better understand how augmentation strategies and protocols change model behavior. 

\textit{Theoretical Analysis.} Extending the feature distortion analysis \cite{Kumar22_FinetuningDistorts} to analytically explain the benefits of hardness-promoting augmentations is also an actively pursued direction.

\bibliography{icml22}
\bibliographystyle{icml2022}

\end{document}